\documentclass[conference]{IEEEtran}
\IEEEoverridecommandlockouts
\usepackage{cite}
\usepackage{amsmath,amssymb,amsfonts}
\usepackage{algorithmic}
\usepackage{graphicx}
\usepackage{textcomp}
\usepackage{xcolor}
\usepackage{tabularx, makecell, array}

\def\BibTeX{{\rm B\kern-.05em{\sc i\kern-.025em b}\kern-.08em
    T\kern-.1667em\lower.7ex\hbox{E}\kern-.125emX}}
\begin{document}

\title{Implicit 3D scene reconstruction using deep learning towards efficient collision understanding in autonomous driving.\\
}

\author{\IEEEauthorblockN{ Akarshani Ramanayake}
\IEEEauthorblockA{\textit{School of Computing} \\
\textit{Informatics Institute of Technology}\\
Colombo, Sri Lanka \\
0009-0000-3581-8900}
\and
\IEEEauthorblockN{Nihal Kodikara}
\IEEEauthorblockA{\textit{School of Computing} \\
\textit{Informatics Institute of Technology}\\
Colombo, Sri Lanka \\
nihal.k@iit.ac.lk}}

\maketitle

\begin{abstract}
In crowded urban environments where traffic is dense, current technologies struggle to oversee tight navigation, but surface-level understanding allows autonomous vehicles to safely assess proximity to surrounding obstacles. 3D or 2D scene mapping of the surrounding objects is an essential task in
addressing the above problem. Despite its importance in dense vehicle
traffic conditions, 3D scene reconstruction of object shapes with
higher boundary level accuracy is not yet entirely considered in
current literature. The sign distance function represents any shape
through parameters that calculate the distance from any point in
space to the closest obstacle surface, making it more efficient in terms of
storage. In recent studies, researchers have started to formulate
problems with Implicit 3D reconstruction methods in the
autonomous driving domain, highlighting the possibility of using
sign distance function to map obstacles effectively. This research
addresses this gap by developing a learning-based 3D scene
reconstruction methodology that leverages LiDAR data and a
deep neural network to build a the static Signed Distance Function
(SDF) maps. Unlike traditional polygonal representations, this
approach has the potential to map 3D obstacle shapes with more
boundary-level details. Our preliminary results demonstrate that
this method would significantly enhance collision detection
performance, particularly in congested and dynamic
environments.
\end{abstract}


\section{Introduction}

Collision detection remains a significant challenge in the
field of autonomous driving due to the complex and dynamic
nature of the real-world environment, where vehicles must
Navigate through a multitude of unpredictable obstacles, varying
weather conditions, and diverse traffic patterns. These factors
necessitate real-time processing and accurate perception of
surrounding objects, requiring advanced algorithms to
distinguish between static and dynamic obstacles, accurately
predict their movements, and assess proximity to avoid
collisions. Furthermore, limitations in the current sensor
Technologies, such as LiDAR and cameras, can lead to issues
with depth perception, occlusion, and detection of blind spots,
compounding the difficulty of ensuring reliable collision
avoidance in diverse driving scenarios.

Current research  has largely focused on vision-based
techniques to perform autonomous vehicle tasks, and one of the
primary challenges of it are obtaining accurate depth information,
which is vital for the vehicle's ability to perceive its surroundings
and plan safe maneuvers. It has also predominantly relied on
the use of polygonal representations, such as bounding boxes, to
define the safe zone of AV from surrounding objects. While this
approaches have proven useful in many scenarios, but it still struggles
in complex environments, particularly during traffic congestion.
In such situations, a lack of boundary-level understanding of
objects can lead to algorithmic failures in collision avoidance
tasks.

This research addresses the limitations of the current collision
avoidance methods by introducing a novel, learning-based
approach to boundary-level 3D scene reconstruction, marking
the first attempt in the autonomous driving research community
to apply such methods in this context. Using the implicit
properties and depth perception capabilities of the learning-based signed distance function (SDF), this work aims to
enhance 3D scene reconstruction, leading to a more accurate
understanding of the environment. Specifically, a deep neural
network is proposed to utilize LiDAR data to construct a static
SDF maps, which provide improved representations of
obstacles, including their shapes and spatial locations. This
continuous nature of the SDF enables the approach to effectively
Overcome the shortcomings of the traditional bounding box
methods in object detection, thereby facilitating more effective
Collision avoidance in congested and complex traffic scenarios.

\subsection{Research questions}
\begin{itemize}
\item How effective is SDF in representing unknown maps for
Collision avoidance in autonomous driving?”

\item Would LIDAR-based SDF 3D reconstruction be
effective for collision avoidance in autonomous driving?

\end{itemize}

\section{RELATED WORKS}

We have structured our literature review to explore three
fundamental aspects of 3D scene mapping in the context of
autonomous driving. First, we investigate the existing shape
map generation methods, focusing on finding existing issues in current methods.
Second, we examine their suitability for real-time, dynamic
applications, and finally, how extensively the current research
community has applied the Signed Distance Function (SDF) within the autonomous driving domain, highlighting their strengths and limitations in enhancing 3D scene understanding.

\subsection{What are existing scene map generation methods in
autonomous driving?}
A symbolic and abstract representation of physical
components, such as roads, buildings, and lakes, is called an
object map. It depicts actual sceneries in two or three
dimensions, complete with object position data and feature
information. Obstacle maps for autonomous vehicles can be
created either in real-time or in advance . While real-time maps
focus primarily on local details, prebuilt maps provide more
comprehensive information on global geometry but struggle
with dynamic object detection. Autonomous vehicle maps
can be classified into two categories. The first includes maps
built from 3D point clouds, which are efficient and closely
aligned with raw sensor data. However, these maps are limited
by their dependence on specific sensors and their instability
under varying environmental conditions. The second category
provides a broader environmental representation, offering
greater interoperability at the cost of sensor efficiency. Map-based a priori localization relies on matching the current sensor
readings with detailed pre-built maps to find the best match \cite{levinson_map-based_2007}.

\subsection{How suitable is the continuous signed distance function for
Learning and representing unknown maps in autonomous
driving?}
Most image-based real-time 3D reconstruction pipelines use
depth map fusion \cite{schops_3d_2015} \cite{yang_real-time_2017}, where single-view depth maps from each keyframe is independently estimated and then fused into
a Truncated Signed Distance Function (TSDF) \cite{hou_multi-view_2019} \cite{liu_neural_2019}. This approach presents challenges, including inconsistent scale factors and inefficiencies from repeatedly estimating the same
surface across multiple keyframes. 

In contrast, \cite{sun_neuralrecon_2021} proposes a framework that reconstructs and fuses 3D geometry directly in a volumetric TSDF representation
using monocular images, leveraging sparse convolutions for
real-time performance. This method eliminates redundant depth
calculations and produces globally coherent 3D scene geometry.
\cite{slavcheva_sdf-2-sdf_2018} proposes a dense 3D reconstruction method from RGB-D data using SDF-2-SDF registration, which improves
reconstruction precision and trajectory estimation without
relying on a pose network. This hybrid system uses the GPU for
real-time tracking and the CPU for optimization, enabling
applications in larger-scale object reconstruction and SLAM.
Although the approach requires significant storage for high-resolution voxel grids, future improvements are expected to enhance performance in more complex environments. The literature review provides evidence that the continuous signed
distance function is suitable for developing dynamic shapes in real-time setups.

\subsection{How implicit 3D reconstruction methods have contributed
to the autonomous domain?}

\cite{sun_neuralrecon_2021} Address the need for high-definition (HD) semantic maps in self-driving vehicles. Maps, built from sensor data like LiDAR and cameras, are crucial for safe navigation, but face challenges in real world
environments, especially in occluded settings and bad weather. The method enhances online map prediction using Neural Map prior to reducing memory consumption through a sparse tile format. \cite{zakharov_autolabeling_2020} Introduces an auto-labelling pipeline for 3D object detection using sparse point clouds and 2D detections. Their system leverages DeepSDF and a novel differentiable renderer to automatically generate labels for training without extensive human annotation. \cite{liu_mv-deepsdf_2023} presents MV-DeepSDF, a framework for 3D vehicle reconstruction using multi-sweep point clouds, improving
fidelity over single-view methods by leveraging complementary information across sweeps. \cite{rao_-vehicle_2021} address in-vehicle object-level 3D reconstruction with a low-cost solution named monocular 3D Shaping, which relies on a single camera frame for automatic parking, showing how
integrating Lidar enhances the occupancy prediction.

\section{SIGN DISTANCE FUNCTION AND CONFIDENCE CALCULATION}

Signed Distance Function (SDF) is a mathematical function that, for any given spatial point in space, computes the distance to the nearest surface of an object, along with a sign. When modeling the sign distance function for an unknown object, the SDF of the object is S(p) for the given spatial point p in R3, returning the corresponding distance to the closest location on the
boundary of the object. Sign specifies whether p is inside (negative) or outside (positive) of the watertight surface.

To calculate the Euclidean distance between the query point and the nearest obstacle point, all obstacle surface points should be iteratively searched to find the minimum distance. Since the brute force method has a high time complexity, this approach becomes inefficient as the number of query points increases, especially if there are many queries about finding the nearest neighbor. Therefore, obstacle points were first structured using a K-D tree algorithm, partitioning the space into 50 leaf nodes. Then, for each point in the final dataset, which included both augmented data and obstacle points, the minimum Euclidean distance was searched using the partitioned space.

Certain points were initially identified as having a negative distance
to an object, but may possess a positive distance from the object's surface because LIDAR ray could only capture the visible surface of the object. So the negative distance becomes higher, and confidence in distance becomes a negative sign becomes lower. This discrepancy highlights the need for a mechanism to gauge the reliability of distance predictions.

\subsection{Requirements when calculating confidence score.}
\begin{itemize}
\item If the sign distance is minus, confidence at the upper
bound of the minus sign distance should be 1. If the sign distance is negative, confidence at the lower bound of the negative sign distance should be near 0. Since confidence is a probability value, confidence should be between 0 and 1.
\item Sometimes objects located far away from the lidar sensor location, due to the limitation of the data augmentation procedure, the object shows high confidence for high minus distances. That minus distance confidence should be very low. The confidence equation should be adaptable according to the maximum minus distance along the ray in the dataset.
\end{itemize}

Building upon the work \cite{camps_learning_2022}, a confidence calculation method given in Eq.1(1) has been adopted. In this method, points with positive distances to the object and those lying on its surface are assigned a confidence value of 1. As the minus distance from the objects increases, the confidence in the prediction gradually decreases.

\begin{equation}
C(p) =
\begin{cases}
1 & \text{if } S(p) \geq 0 \\
\frac{b^{\omega - 1}}{b - 1} + 1e^{-7} & \text{otherwise,}
\end{cases}
\end{equation}
Where $\omega = 1 - \frac{d(p)}{d_{max}}$,is a normalized distance between 0 and 1, 

$d(p)$ is a distance away from the surface along the ray $d_{max}$ is the maximum distance along the ray. b is the hyperparameter.

\section{METHODOLOGY}

We assume the AV with pose at a discrete time instance knows its own pose, but it does not know the surrounding obstacles. However, the AV is equipped with a LiDAR sensor with which it can sample points on the surfaces of obstacles.
The primary objective of this study is to explore the potential of 3D scene reconstruction of surrounding obstacles, specifically through the use of LIDAR point clouds. In this research, static scenes were loaded from the NuScene dataset \cite{caesar_nuscenes_2020}, and as discussed in section V, pre-processing was done for the loaded scene.

After pre-processing, the model learns the obstacle shape using a 3-layer fully connected neural network with tanh, Huber loss, Adam, 0.4 as activation function, loss function, optimizer, and learning rate as a sign distance field. Fourier feature encoding helps the network capture fine details
by transforming spatial inputs into a higher frequency space. Therefore, before hidden layers, A Fourier feature encoding layer was added after the input layer
that takes a spatial input point in $\mathbb{R}^3$ and returns a frequency
vector output in $\mathbb{R}^{64}$. The networks then output two values, the sine distance along with a confidence value. The Decision boundary of the neural
network, where sign distance s(p) = 0 considered the 3D shape of the obstacle.

\section{PRE-PROCESSING DATA FOR DATA INGESTION}
The NuScenes dataset, developed by Motional, contains LiDAR scans, and each LiDAR scan timestamp corresponds to the completion of a full sensor rotation. The dataset provides points from 32 classes, consisting of 23 foreground classes (e.g., vehicles,
pedestrians) and 9 background classes (e.g., roads, sidewalks).
During the points filtering process, background classes and the ego
vehicle class was removed.

After filtering data points, the Directed Hausdorff distance was
used to measure the similarity between consecutive static point
cloud. If the distance exceeded a predefined threshold, the static
point cloud was passed to the next process. The passed point
cloud is now filtered using the ground filtering process of treating
points with a Z-coordinate less than -1.563 meters (the vehicle’s
height) as floor points, while all other points were considered
potential obstacles. 

To facilitate the comprehensive learning of signed distance
fields (SDF) using the SDF ANN algorithm, it is essential to
include points representing both positive and negative distances
from the surface of objects. However, conventional lidar point
clouds typically capture points only from the exposed surface of
objects, neglecting those that lie within or beyond the surface.
This research has tested two data augmentation strategies to
augment positive and negative points to the final points set.

To model a neural network, it is important to have both
independent variables and dependent variables. Therefore,
respective sign distance values and confidence values were added
for each obstacle point.

\section{EXPERIMENTS SETUP}
\subsection{Data augmentation techniques}
In this study, LiDAR point cloud data from the NuScenes
dataset was used to facilitate the Signed Distance Function (SDF)
reconstruction. The data augmentation process addresses the
challenge of creating balanced positive and negative sample
points for effective learning. Positive points represent the
distance from the LiDAR origin to the object's surface, while
negative points extend beyond the object. To avoid the issue of
infinite negative distances, which can disrupt the learning
process, negative points were truncated. This approach ensures
more accurate and stable training data for 3D scene
reconstruction. in this research, two sampling methods were tested.

Uniform Sampling: Fig. 1. In the initial approach, positive and
negative points were sampled uniformly along each LiDAR ray.
Positive points were taken from the LiDAR origin to the object
surface, while negative points were generated beyond the
termination point. This uniform sampling led to a concentration
of positive points in close proximity to the sensor, while
negative points spread across a wide range of distances, leading
to an imbalance in data distribution.

Gaussian Sampling: Fig. 3. To address the imbalance and reduce
the number of distant negative points, Gaussian sampling was
introduced near the LiDAR termination point. This method
ensured a more balanced distribution of negative and positive
points, especially by reducing the frequency of very high
negative distance points. This adjustment is crucial, as excessive
negative sampling at a far distance leads to inaccuracies and
degrades the model's performance.

\subsection{Testing the model loss compared to model size.
}
Since the effective complexity of the model to build the
scene is unknown, and a large network is too slow to train online,
The main purpose of this test setup is to identify the impact of
the model accuracy with respect to model size. The hypothesis of the experiment is “ When the number of parameters increases along with skip connections, model accuracy increase”. In this scientific experiment, the number of layers is taken as a dynamic parameter, while all other parameters like activation functions, loss functions, the number of nodes in a single layer, and optimizer were taken as constant variables relatively, Relu, Huber loss, 64, Adam. To increase model parameters while keeping a constant dimension, a linear layer with input dimension 1x64 and output dimension 1x64 without skip connections, as shown in Fig. 6, and with skip connections as shown in Fig. 5, altogether 40 models were used.

\subsection{Model performance vs Fourier feature}
In most scenarios, adding Fourier feature layers improves the
model convergence. In this experiment, 3 models were used. Model 1 is a general ANN algorithm injected with the uniformly augmented coordinate set. Model 2 is a general ANN algorithm injected with a point cloud sampled using the Gaussian distribution-based augmentation method. Model 3 is a general ANN algorithm that has a Fourier feature encoder injected with a point cloud sampled using the Gaussian distribution-based augmentation method. 

\section{EXPERIMENT RESULTS}
\subsection{Data augmentation techniques}
Figures 2 and 4 illustrate the Signed Distance Function (SDF) and confidence predictions generated by the neural network after applying uniform and Gaussian distributed data augmentations. Several key observations were made during the experiment. For positive SDF values—indicating points outside the object—the predicted confidence consistently falls within an error margin of ±0.5 around the expected confidence value of 1. However, for negative SDF values representing points inside the object, the model produced invalid confidence scores. In the uniformly sampled dataset, some negative SDF predictions resulted in confidence values ranging from 0 to -3, which are clearly invalid. Similarly, in the Gaussian distributed dataset, most invalid predictions for negative SDF values fell within the range of 0 to -0.5. As shown in Table 1, despite these augmentation efforts, the neural network models did not demonstrate the expected performance improvements.

\subsection{Testing the model loss compared to model size.}

According to Fig. 6, increasing the number of hidden layers does not lead to improved model performance. While skip connections begin to show an effect starting from the 18th layer, the overall performance remains suboptimal. Interestingly, models with fewer layers were able to achieve better performance, suggesting that a simpler architecture is more effective for this specific task.

\subsection{Model performances with Fourier features.}

Low-frequency features in the spectral domain were effectively captured by incorporating a Fourier feature encoder after the input layer. As shown in TABLE 1, this addition significantly enhanced the model’s performance, nearly doubling it, highlighting the encoder’s ability to improve feature representation and learning efficiency.

\begin{table}
\begin{tabularx}{\textwidth}{>{\centering\arraybackslash}p{25mm} >{\centering\arraybackslash}p{23mm}  >{\centering\arraybackslash}p{23mm}}

\rule{82mm}{0.5pt} \\ 
\textbf{ Model} & \textbf{Augmentation} & \textbf{Huber Loss} \\ 
\rule{82mm}{0.5pt} \\

ANN & UNIFORM & 0.4 \\
ANN & GAUSSIAN & 0.37 \\
\makecell{ANN  \\+ FF ENCODER} & GAUSSIAN & 0.18 \\
\rule{82mm}{0.5pt} \\ 

\end{tabularx}
\caption{Huber loss comparison with Fourier feature encoder and data augmentation techniques}
\end{table}

\begin{figure}[htbp]
 \includegraphics[width=\linewidth]{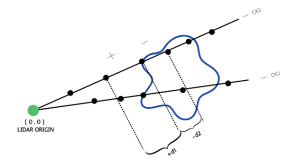}
        \caption{Uniform distribution to get positive and negative points during data augmentation.}
        \label{fig:uniform_dist}
\end{figure}

\begin{figure}[htbp]
 \includegraphics[width=\linewidth]{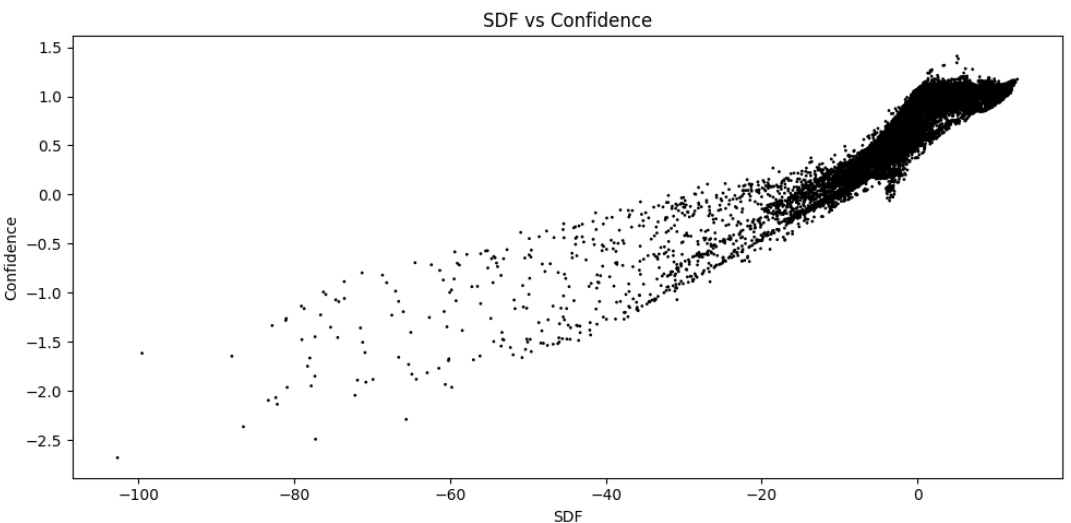}
        \caption{ Confidence vs Sign distance value graph of the final prediction after using a uniformly distributed data augmentation method.}
        \label{fig:uniform_confidence_sdf}
\end{figure}

\begin{figure}[htbp]
 \includegraphics[width=\linewidth]{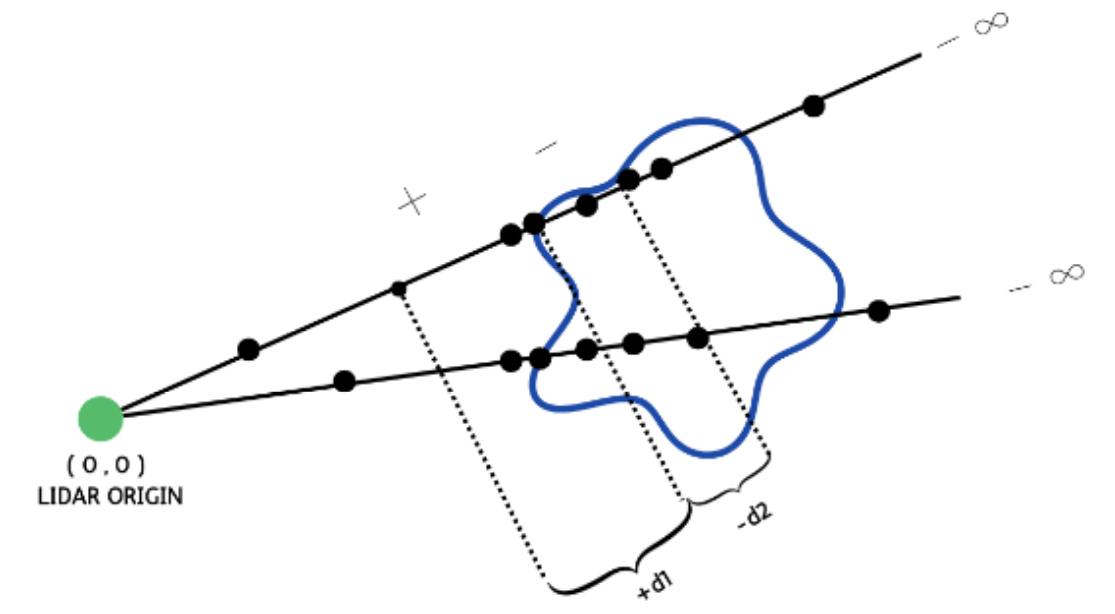}
        \caption{Gaussian distribution to get positive and negative points during data augmentation.}
        \label{fig:gaussian_sampling}
\end{figure}

\begin{figure}[htbp]
 \includegraphics[width=\linewidth]{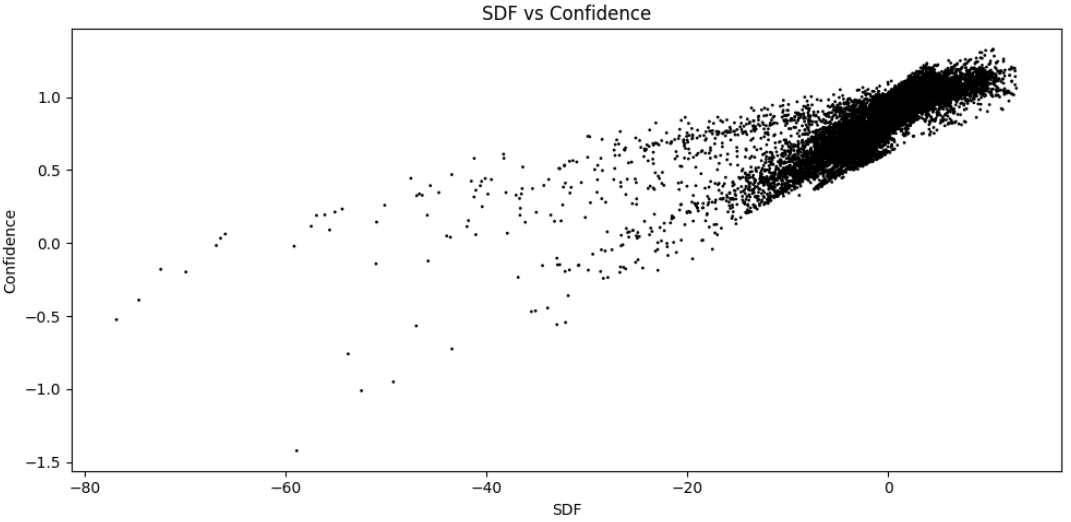}
        \caption{ Confidence vs Sign distance value graph of the final prediction after using a Gaussian distributed data augmentation method.}
        \label{fig:gaussian_confidence_sdf}
\end{figure}

\begin{figure}[htbp]
 \includegraphics[width=\linewidth]{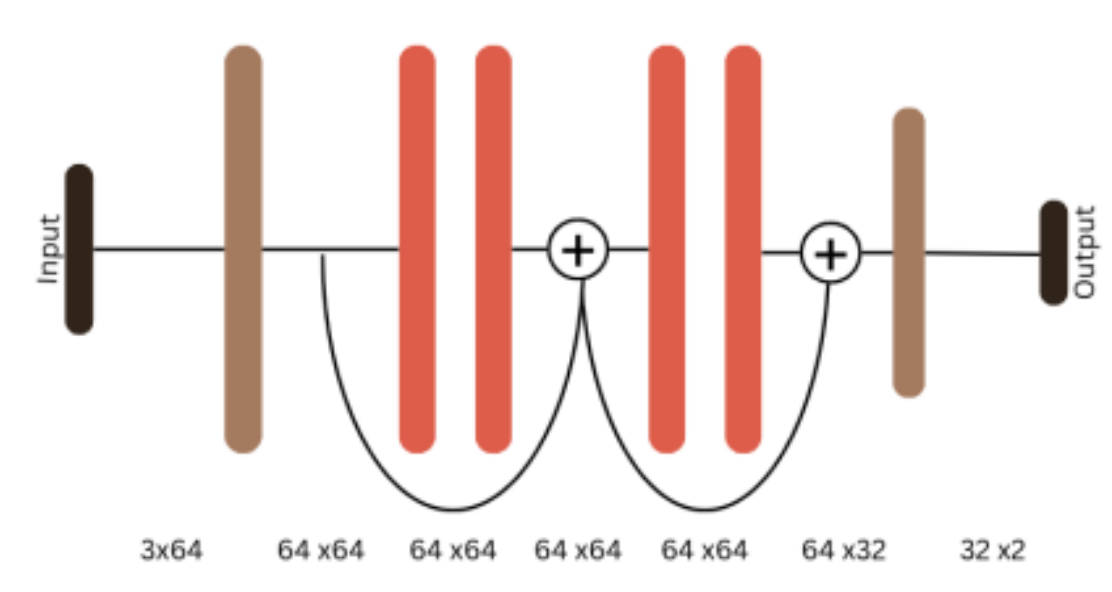}
        \caption{The architecture tested with skip connections.}
        \label{fig:skip_connections_arch}
\end{figure}

\begin{figure}[htbp]
 \includegraphics[width=\linewidth]{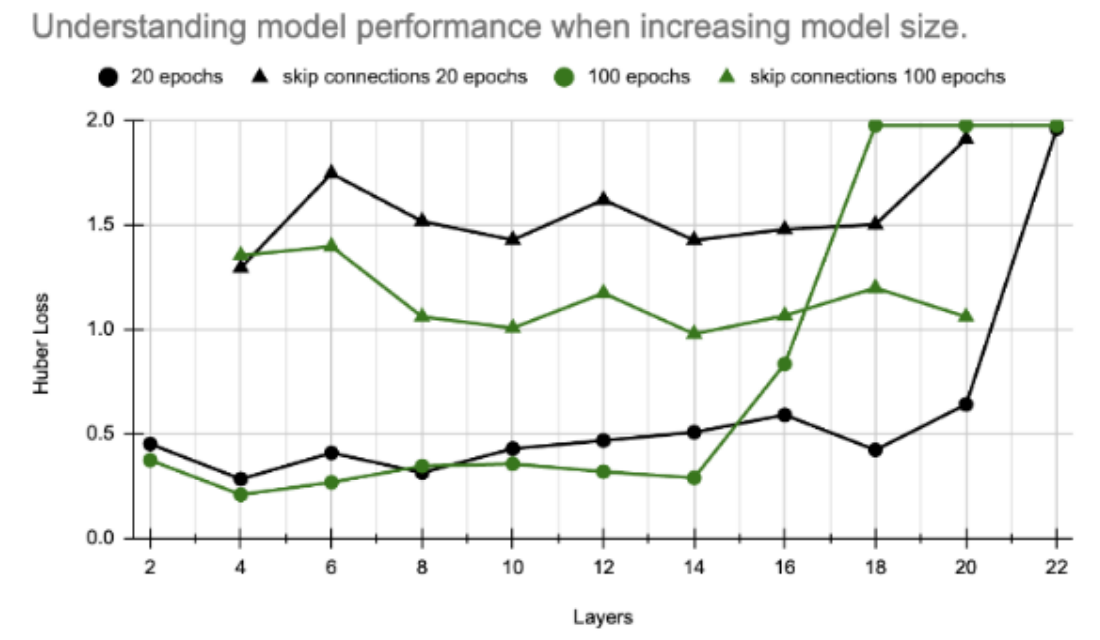}
        \caption{Test loss comparison of neural networks ingested with different numbers of trainable parameters.}
        \label{fig:test_loss}
\end{figure}

\begin{figure}[htbp]
 \includegraphics[width=\linewidth]{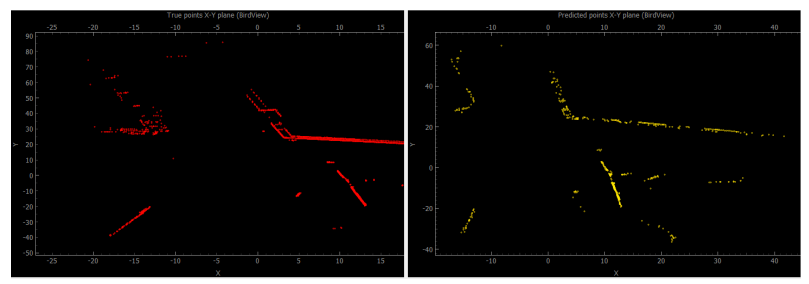}
        \caption{Bird view of ground truth and predicted scene, respectively.}
        \label{fig:bird_view}
\end{figure}

\section{LIMITATIONS}
\subsection{Limitations of data augmentation techniques.}

Positive sample points in LiDAR data can be affected by interruptions from nearby objects, especially when the detected object is farther away. This can lead to inaccuracies in distance estimation, reducing the model's accuracy with increasing distance from the sensor.

\subsection{Limitations of point sampling using LIDAR sensor.}

LiDAR sensors lose accuracy at greater distances due to reduced point density caused by limited angular resolution. This results in less detail and degraded performance when capturing distant objects.

\subsection{Limitations of point sampling using LIDAR sensor.}

3D scenes typically include both large obstacles and smaller ones. However, small objects are represented by relatively few points in the point cloud, leading the neural network to treat them as shallow features. Additionally, due to the unpredictable complexity of each scene, the proposed neural network may struggle to adapt effectively, limiting its ability to generalize and learn from varying environments.

\section{CONCLUSION}

In this study, we introduced a methodology for 3D scene
reconstruction of obstacles using LiDAR point clouds and a
neural network-based Signed Distance Function (SDF)
approach. The integration of Fourier feature encoding has
proven to be effective in enhancing the accuracy of the
reconstructed 3D shapes. By addressing the key challenge of data
imbalance, data augmentation techniques were proposed. Our
method allows for more accurate modeling of obstacle surfaces.
The results of this research suggest that learning-based 3D
reconstruction techniques, specifically those employing SDFs,
can provide superior boundary-level accuracy for obstacle
detection. This approach holds great potential for enhancing the
navigation and safety of autonomous vehicles in complex and
densely populated urban environments. Future work could focus
on expanding this model to handle dynamic scenes, which
would further improve its real-world applicability in
autonomous driving systems.

\bibliography{conference_101719}{}
\bibliographystyle{IEEEtran}

\end{document}